\documentclass[runningheads]{llncs}
\usepackage[T1]{fontenc}
\usepackage{graphicx}

\usepackage{amsmath}
\usepackage{amssymb}
\usepackage{float}
\usepackage{multirow}
\usepackage{booktabs}

\usepackage{hyperref}    

\usepackage{color}

\usepackage{cuted}  
\usepackage{etoolbox}
\usepackage{caption}
\usepackage{hypcap} 

\begin{document}
\title{Learning Disentangled Speech- and Expression-Driven Blendshapes for 3D Face Animation}
\titlerunning{Learning Disentangled Speech- and Expression-Driven Blendshapes}
%

\author{Yuxiang Mao\inst{1,2} \and
Zhijie Zhang\inst{2} \and
Zhiheng Zhang\inst{1,2} \and
Jiawei Liu\inst{3} \and
Chen Zeng\inst{3} \and
Shihong Xia\inst{1,2}\thanks{corresponding author, xsh@ict.ac.cn}}
\authorrunning{Y. Mao et al.}
\institute{Institute of Computing Technology, Chinese Academy of Sciences\\
\email{\{maoyuxiang22z,zhangzhiheng20g,xsh\}@ict.ac.cn}
\and
University of Chinese Academy of Sciences
\and
Huadian (Beijing) Co-Generation Co., Ltd.}
%
\maketitle              
\begin{abstract}
Expressions are fundamental to conveying human emotions. With the rapid advancement of AI-generated content (AIGC), realistic and expressive 3D facial animation has become increasingly crucial. Despite recent progress in speech-driven lip-sync for talking-face animation, generating emotionally expressive talking faces remains underexplored. A major obstacle is the scarcity of real emotional 3D talking-face datasets due to the high cost of data capture. To address this, we model facial animation driven by both speech and emotion as a linear additive problem. Leveraging a 3D talking-face dataset with neutral expressions (VOCAset) and a dataset of 3D expression sequences (Florence4D), we jointly learn a set of blendshapes driven by speech and emotion. We introduce a sparsity constraint loss to encourage disentanglement between the two types of blendshapes while allowing the model to capture inherent secondary cross-domain deformations present in the training data. The learned blendshapes can be further mapped to the expression and jaw pose parameters of the FLAME model, enabling the animation of 3D Gaussian avatars. Qualitative and quantitative experiments demonstrate that our method naturally generates talking faces with specified expressions while maintaining accurate lip synchronization. Perceptual studies further show that our approach achieves superior emotional expressivity compared to existing methods, without compromising lip-sync quality.


\end{abstract}
\section{Introduction}
\label{sec:intro}
In recent years, speech-driven 3D facial animation has found wide applications in fields such as entertainment~\cite{chen2009analysis}, XR~\cite{li2017learning}, and video conferencing~\cite{wang2021oneshot}. This technology enables the generation of high-quality 3D talking faces from arbitrary speech inputs. With the rapid advancement of deep learning, the field has achieved remarkable progress~\cite{cudeiro2019capture,danecek2023emotional,fan2022faceformer,richard2021meshtalk,thambiraja2023imitator,xing2023codetalker,shen2023difftalk,zhou2020makelttalk,peng2023emotalk,nocentini2025emovoca,he2023speech4mesh,sun2024diffposetalk}. Recent studies~\cite{danecek2023emotional,peng2023emotalk,nocentini2025emovoca,kim2024laughtalk,karras2017audio} have extended beyond lip synchronization, emphasizing the role of facial expressions as crucial non-verbal cues.

The core challenge in emotional talking-face animation lies in accurately conveying emotion while ensuring that lip motion remains synchronized with the input audio. However, in 3D facial animation, datasets that contain speech-driven 3D talking faces with diverse emotional expressions are scarce, making relevant research difficult. Existing methods~\cite{danecek2023emotional,peng2023emotalk,he2023speech4mesh,kim2024laughtalk,sun2024diffposetalk} often reconstruct 3D pseudo ground-truth data from emotional 2D video datasets~\cite{wang2020mead,zhang2021flow,livingstone2018ravdess} to train emotion-enhanced speech-driven facial animation models. Yet, such pseudo ground-truth data falls short of real 3D scans in quality, leading to decreased lip-sync accuracy~\cite{nocentini2025emovoca}, and making it difficult to learn well-disentangled and accurate facial deformations driven by both speech and emotion. We observe that while no single dataset simultaneously contains both factors, real 3D scan datasets for each individual factor are accessible~\cite{cudeiro2019capture,principi2023florence}. This motivates us to adopt a data-driven approach to learn realistic facial motion patterns from these high-quality 3D scan datasets.

\begin{figure}[!t]
\centering
\includegraphics[width=0.9\textwidth]{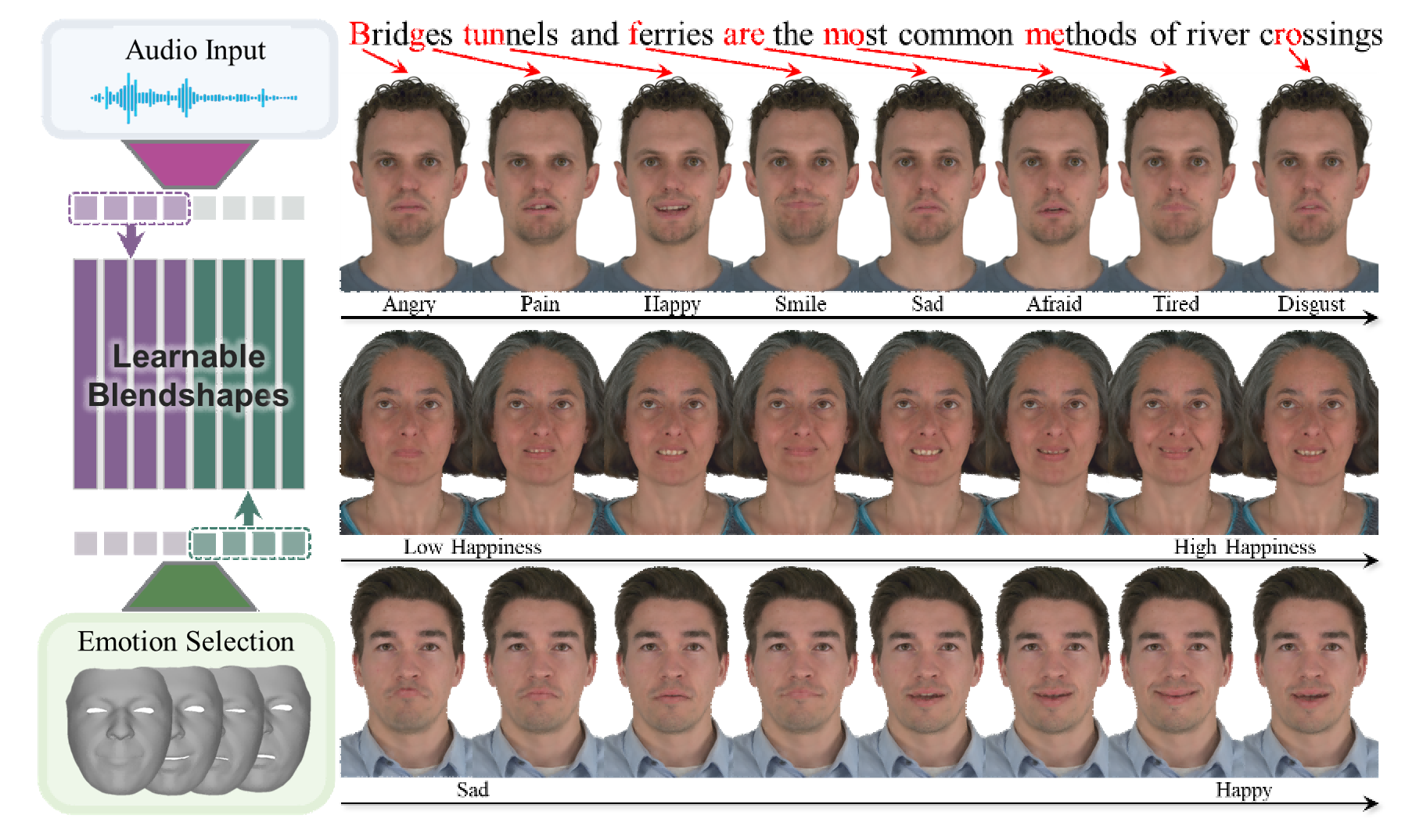}
\caption{We regress the weights of the learned speech- and expression-driven blendshapes respectively, remove the secondary cross-domain deformations, and then combine them to produce realistic, emotionally expressive talking avatars.}
\label{fig:cover}
\vspace{-8pt}
\end{figure}

Blendshape~\cite{chuang2002performance} is a widely used technique in 3D facial animation~\cite{DBLP:journals/jvca/LiuMXYW08}, representing facial motion as a linear combination of a neutral base mesh and multiple predefined deformation targets weighted by adjustable coefficients to generate complex expressions and lip movements. We model speech- and expression-driven animation as a linear additive problem, learning a set of disentangled speech- and expression-driven blendshapes from two real scan datasets: the audio-driven 3D facial animation dataset VOCAset~\cite{cudeiro2019capture} and the 3D facial expression dataset Florence4D~\cite{principi2023florence}. These blendshapes naturally embed emotional deformations into talking faces and can be further mapped to the parameter space of FLAME~\cite{li2017learning}, enabling the animation of Gaussian-based head avatars~\cite{qian2024gaussianavatars}.

However, the training data in each domain inevitably contains certain deformations that originate from the other factor. In VOCAset, for instance, speakers seldom maintain a perfectly neutral expression during speech; in Florence4D, emotional expressions often affect lip articulation. We refer to these interfering components as \textit{secondary deformations}—domain-specific deformations that should ideally be attributed to the other domain. Due to the presence of such secondary deformations, directly combining speech- and expression-driven deformations can lead to artifacts, such as the lips failing to fully close (Fig.~\ref{fig:ablat}). 

To address this, we introduce a sparsity constraint loss that encourages the disentanglement of speech and expression factors while still allowing the model to account for secondary deformations during training. During inference, our deformation fusion module regresses blendshape weights for speech and expression deformations respectively, removes the cross-domain components, and then combines the two to synthesize talking faces with accurate lip synchronization and emotional expressiveness (Fig.~\ref{fig:infer}).

In summary, the main contributions of our work are as follows:
\begin{itemize}
    \item We present a data-driven approach to learn disentangled speech- and expression-driven blendshapes from real 3D face scan datasets, enabling the capture of realistic lip motion and emotional facial deformations with high fidelity.
    \item We introduce a sparsity constraint loss that removes secondary cross-domain deformations, achieving clean separation between speech and expression effects and preventing artifacts when combining the two factors.
    \item The learned blendshape coefficients are mapped to the FLAME parameter space, enabling our deformation fusion module to further animate Gaussian head avatars in real-time, producing natural, emotionally expressive 3D talking faces with accurate lip synchronization.
\end{itemize}

\section{Related Work}

Currently, a number of efforts~\cite{cudeiro2019capture,danecek2023emotional,fan2022faceformer,richard2021meshtalk,thambiraja2023imitator,xing2023codetalker,shen2023difftalk,zhou2020makelttalk,peng2023emotalk,nocentini2025emovoca,he2023speech4mesh} have developed methods for generating speech-driven 3D talking faces.

VOCA~\cite{cudeiro2019capture} presents a speaker-independent method for 3D facial animation that captures diverse speaking styles. FaceFormer~\cite{fan2022faceformer} is the first to adopt a transformer-based model that autoregressively produces facial motion from vocal inputs. CodeTalker~\cite{xing2023codetalker} develops a discrete codebook for general facial movements and employs a framework similar to FaceFormer to synthesize these animations. However, these approaches mainly emphasize lip movements while overlooking the subtle non-verbal cues that occur during speech.

Karras et al.~\cite{karras2017audio} incorporates a trainable emotional state vector, assigning one vector to each training sample to construct an emotion database. However, the learned vectors lack explicit semantic interpretation, and the architecture does not include a disentanglement mechanism to separate the influence of speech and emotion signals. MeshTalk~\cite{richard2021meshtalk} decouples voice and non-voice factors by introducing a novel cross-modality loss, encouraging the model to learn upper-face movements independent of audio and accurate mouth movements that depend solely on audio. However, emotional cues can also manifest in the lower face region, such as in the corners of the mouth, making the assumption of strict upper–lower face disentanglement potentially too strong.

Currently, to the best of our knowledge, there is no publicly available real 3D scan dataset for talking-face animation with diverse emotions. Several studies~\cite{danecek2023emotional,peng2023emotalk,he2023speech4mesh,kim2024laughtalk,sun2024diffposetalk} extract pseudo ground-truth 3D data from emotional 2D video datasets such as MEAD~\cite{wang2020mead}, HDTF~\cite{zhang2021flow}, and RAVDESS~\cite{livingstone2018ravdess}. However, experiments~\cite{nocentini2025emovoca} show that this reconstructed data still falls short of real scan data in terms of geometric detail and lip accuracy.  

EmoVOCA~\cite{nocentini2025emovoca} leverages the widely used real 3D scan dataset with paired audio (VOCAset~\cite{cudeiro2019capture}) and the 3D facial expression dataset (Florence4D~\cite{principi2023florence}), synthesizing a new 3D dataset that combines the effects of speech and expressions on facial deformations via a double-encoder-shared-decoder architecture. However, this method assumes that the features in the two datasets are perfectly disentangled into speech and expression deformations, which differs from our view. We argue that both datasets inevitably contain subtle deformations attributable to the other domain, which should be disentangled during training.

Unlike the above methods, our approach learns a set of explicitly disentangled blendshapes for the two driving factors from VOCAset and Florence4D. By applying a sparsity constraint loss to the regressed blendshape weights, we separate the secondary cross-domain deformations present in each dataset, thereby achieving better disentanglement.

\section{Method}
Our method involves first training a speech-driven facial animation model to generate speech-driven deformations from the input audio. Then, our deformation fusion module regresses a set of learned blendshape weights for both the speech and specified expression (sequence) deformations respectively. By removing cross-domain artifacts and combining these two sets of weights, we achieve natural and realistic emotional talking face animation.

\begin{figure}[!t]
\centering
\includegraphics[width=\textwidth, ]{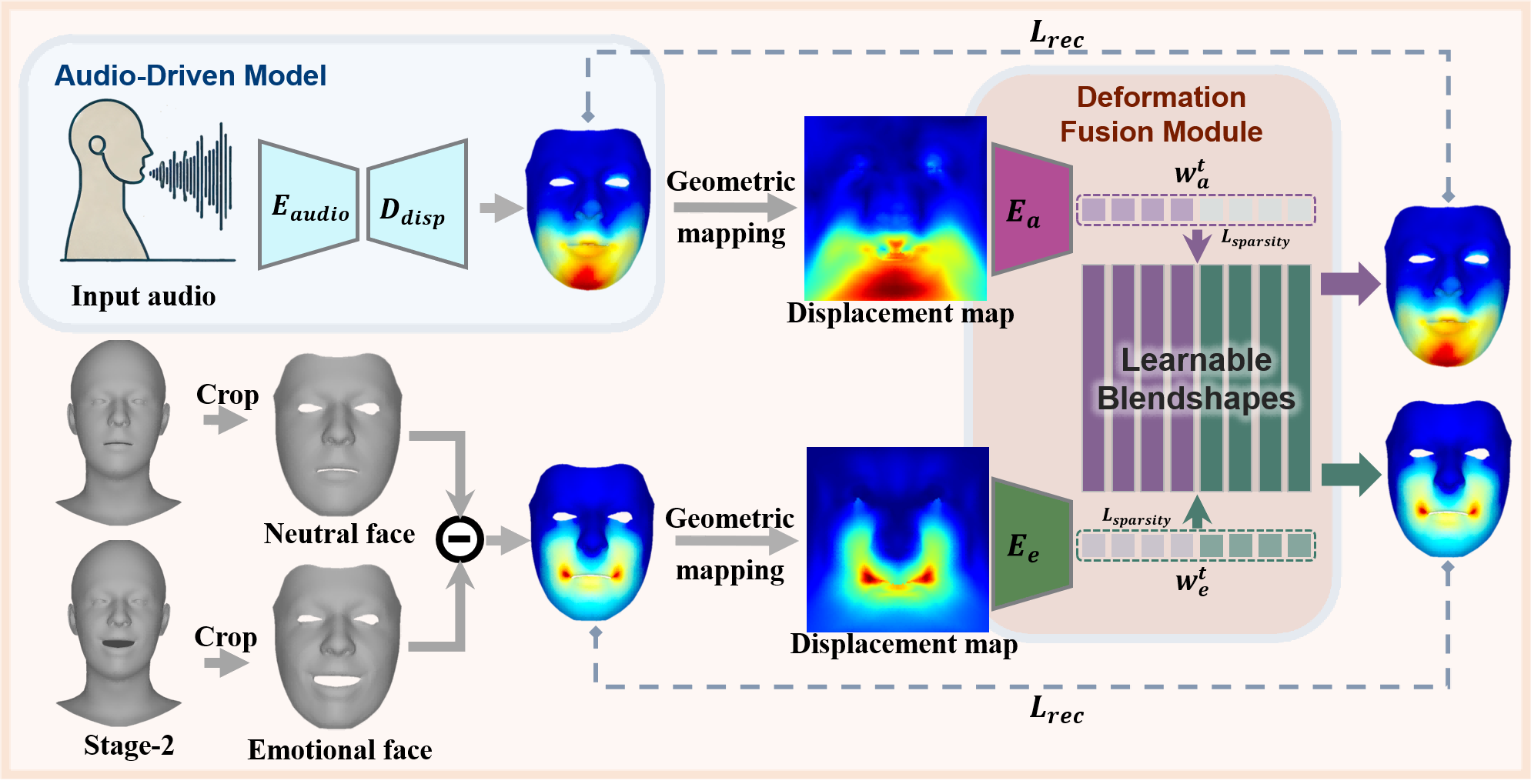}
\caption{\textbf{Architecture.} Our model comprises two stages. In the first stage, an audio-driven talking face model is employed to generate speech deformations from the input audio. In the second stage, we jointly learn the blendshapes for speech- and expression-driven deformations, with the encoders $E_a$ and $E_e$ trained simultaneously to regress their corresponding weights.}
\label{fig:arch}
\vspace{-8pt}
\end{figure}

\subsection{Data Preparation and Geometric Mapping}
Unlike point clouds, mesh models have explicit topology that encodes neighborhood relationships in 3D space. However, representing mesh vertex coordinates as simple column vectors fails to preserve this local connectivity, and is not naturally compatible with CNN architectures, which are designed to exploit such spatial locality. To address this, we adopt the method of Fan et al.~\cite{fan2023unpaired}, which represents 3D shapes on a square image grid while preserving the adjacency relationships of 3D vertices.

Specifically, for a speech-driven mesh $M_a$ from VOCAset or a expression-driven mesh $M_e$ from Florence4D, we first crop the frontal face region. Following the procedure of Fan et al.~\cite{fan2023unpaired}, we apply a dense registration method~\cite{fan2023towards} to align them with the provided face template, resulting in meshes with unified topology, denoted as $V_a \in \Re ^ {n_v \times 3}$ and $V_e \in \Re ^ {n_v \times 3}$, where $n_v = 10857$. We then subtract their corresponding canonical face to obtain per-vertex displacements $\Delta V_a$ and $\Delta V_e$ representing speech- and expression-driven deformations. As shown in Fig.~\ref{fig:arch}, these displacements are subsequently mapped, via interpolation, to geometry displacement maps $D_a \in \Re ^ {H \times W \times 3}$ and $D_e \in \Re ^ {H \times W \times 3}$ on a square image grid~\cite{fan2023unpaired}:

\begin{equation}
D_{i,j}={w_{i,j,1} V_{I_{i,j,1}} + w_{i,j,2} V_{I_{i,j,2}} + w_{i,j,3} V_{I_{i,j,3}}},
\end{equation}

where $I_{i,j,*}$ and $w_{i,j,*}$ denote the indices of the three corresponding vertices in $V$ and their barycentric interpolation weights for each vertex (pixel) $D_{i,j}$ in the resulting square image grid.

\subsection{Audio-Driven Model}
First, we construct a speech-driven model to generate corresponding mesh deformations from input audio. Following FaceFormer~\cite{fan2022faceformer}, we train this model on the registered VOCAset meshes $\{V_a\}$. The audio encoder $E_{\text{audio}}$ in FaceFormer adopts the wav2vec~2.0 model~\cite{baevski2020wav2vec}, which includes an audio feature extractor based on a Temporal Convolutional Network (TCN)~\cite{lea2016temporal,lea2017temporal,lea2017tcn} to transform raw audio into features, which are then processed by a Transformer encoder to produce higher-level speech representations. The decoder $D_{\text{disp}}$ leverages a causal self-attention mechanism to learn temporal motion representations, along with a cross-modal attention mechanism to align the audio and motion modalities, thereby achieving accurate synchronization. This process of predicting the current-frame deformation based on the input audio $A^{1:T}$ and the previous deformations $\Delta \hat{V}_{A}^{1:t-1}$ can be formulated as:
\begin{equation}
\Delta \hat{V}_{A}^t = D_{\text{disp}}\left(E_{\text{audio}}(A^{1:T}), \Delta \hat{V}_{A}^{1:t-1}\right).
\end{equation}

\subsection{Learning Speech- and Expression-Driven Blendshapes}
\subsubsection{Formulation}
We model speech and expression as two additive driving factors in a linearly additive problem. By linearity, the resulting deformation can be formulated as:
\begin{equation}
\Delta V = f_A(A) + f_E(E) = B_A \vec{w_A} + B_E \vec{w_E},
\end{equation}
where $B_A$ and $B_E$ are the blendshapes for speech and expression, and $w_A$ and $w_E$ are the corresponding weights.

Given speech- and expression-driven deformation data $\Delta V_a$ and $\Delta V_e$ (or $D_a$ and $D_e$), our goal is to learn the blendshape sets $B_A$ and $B_E$. We assume that each domain’s deformation data inevitably contains secondary deformations from the other domain. For example, for an expression-driven deformation $\Delta V_{E_0}$ in the training set, where $E_0$ denotes an arbitrary expression, we have:
\begin{equation}
\Delta V_{E_0} = B_E \vec{w_{E_0}} + B_A \vec{\epsilon},
\end{equation}
where $\vec{\epsilon}$ represents the non-negligible secondary speech-related deformation.

If $\vec{\epsilon}$ is ignored, then for a speech input $A_0$ and expression $E_0$ we have:
\begin{equation}
\Delta V_{A_0} = B_A w_{A_0} + B_E \cdot \vec{0}, \quad
\Delta V_{E_0} = B_A \cdot \vec{0} + B_E w_{E_0}.
\end{equation}
The resulting combined deformation for the target talking face with expression would be:
\begin{equation}
\Delta V_{O_0} = B_A \vec{w_{A_0}} + B_E \vec{w_{E_0}} = \Delta V_{A_0} + \Delta V_{E_0},
\end{equation}
which degenerates into a naive sum (or interpolation) of the two input deformations. 

Our ablation study demonstrates that such naive composition degrades lip articulation, particularly for closed-mouth phonemes, because certain expression deformations introduce undesired mouth openings that are not compensated for. Therefore, $\vec{\epsilon}$ should also be regressed and removed during the fusion of speech and expression deformations. 

Specifically, for the speech-driven deformation $\Delta V_{A_0}$ and the expression-driven deformation $\Delta V_{E_0}$, we have:
\begin{equation}
\Delta V_{A_0} = B_A \vec{w_{A_0}} + B_E \vec{\epsilon_{A_0}}, \quad 
\Delta V_{E_0} = B_A \vec{\epsilon_{E_0}} + B_E \vec{w_{E_0}}.
\end{equation}
The fused output deformation is then given by:
\begin{equation}
\Delta V_{O_0} = B_A w_{A_0} + B_E w_{E_0} = (\Delta V_{A_0} - B_E \vec{\epsilon_{A_0}}) + (\Delta V_{E_0} - B_A \vec{\epsilon_{E_0}}).
\end{equation}

\subsubsection{Architecture}

We jointly learn the blendshapes for speech- and expression-driven deformations within our \textbf{deformation fusion module}, which consists of two independent CNN encoders, $E_a$ and $E_e$, and a linear layer as the decoder. The weight matrix of this linear decoder serves as the learnable blendshapes, while the latent vectors encoded by the encoders serve as the blendshape weights. We partition the blendshapes such that the first half $B_A$ corresponds to speech-driven deformations and the second half $B_E$ corresponds to expression-driven deformations, as illustrated in Fig.~\ref{fig:arch}.

Specifically, in our implementation, each encoder consists of six blocks of convolutional layers, instance normalization layers, and ReLU activation functions. The input speech-driven deformation $\Delta V_{A_0}$ and expression-driven deformation $\Delta V_{E_0}$ are first interpolated to geometry displacement maps $D_{A_0}$ and $D_{E_0}$, respectively. These are then encoded by their corresponding encoders $\mathrm{E}_a$ and $\mathrm{E}_e$ into blendshape weights, which are subsequently passed through the linear decoder to produce the predicted per-vertex displacements $\Delta \hat{V}_{A_0}$ and $\Delta \hat{V}_{E_0}$. By supervising $\Delta \hat{V}_{A_0}$ and $\Delta \hat{V}_{E_0}$ to match the input $\Delta V_{A_0}$ and $\Delta V_{E_0}$, we self-supervise the regression of both the learnable blendshapes and their corresponding weights.

\subsubsection{Sparsity Constraint Loss}

To encourage disentanglement between the learned speech- and expression-driven blendshapes, we introduce a sparsity constraint loss to suppress the cross-domain blendshape weights produced by the encoders. Specifically, the output of the speech encoder $\mathrm{E}_a$ is split into two halves, $\vec{w_{A_0}}$ and $\vec{\epsilon_{A_0}}$, where $\vec{w_{A_0}}$ corresponds to the weights of speech-driven blendshapes $B_A$, and $\vec{\epsilon_{A_0}}$ corresponds to the weights of expression-driven blendshapes $B_E$. We impose a sparsity constraint on $\vec{\epsilon_{A_0}}$ to encourage its values to be small. Similarly, the output of the expression encoder $\mathrm{E}_e$ is split into $\vec{w_{E_0}}$ and $\vec{\epsilon_{E_0}}$, where $\vec{\epsilon_{E_0}}$ corresponds to the weights of speech-driven blendshapes $B_A$. A sparsity constraint is also applied to $\vec{\epsilon_{E_0}}$ to minimize the cross-domain influence:

\begin{equation}
\mathcal{L}_{\text{sparsity}} = \| \epsilon_{A_0} \|_1 + \| \epsilon_{E_0} \|_1,
\end{equation}

It is worth noting that this loss does not enforce $\vec{\epsilon_{A_0}}$ and $\vec{\epsilon_{E_0}}$ to be exactly zero, but allows them to take small nonzero values.

\subsubsection{Overall Losses}

We jointly train the speech encoder $\mathrm{E}_a$, the expression encoder $\mathrm{E}_e$, and the blendshape bases $B_A$ and $B_E$ on the speech- and expression-driven datasets VOCAset and Florence4D. In total, we optimize:
\begin{equation}
\mathcal{L}_{\text{overall}} = \lambda_{\text{rec}}\mathcal{L}_{\text{rec}} + \lambda_{\text{sparsity}}\mathcal{L}_{\text{sparsity}} + \lambda_{\text{Laplace}}\mathcal{L}_{\text{Laplace}} + \lambda_{\text{reg}}\mathcal{L}_{\text{reg}},
\end{equation}
where $\mathcal{L}_{\text{reg}}$ is an $\ell_2$ regularization loss applied to the output deformations, and:
\begin{equation}
\mathcal{L}_{\text{rec}} = \frac{1}{n_v} \sum_{k=1}^{n_v} \omega_k \| \Delta \hat{V}_{k} - \Delta V_{k} \|_2
\end{equation}
is the reconstruction loss, which minimizes the weighted mean squared error between the predicted deformation and the ground truth.

The Laplace smoothing term is defined as:
\begin{equation}
\mathcal{L}_{\text{Laplace}} = \sum_{i=1}^{n_v} \left\| \hat{V}_i - \frac{1}{\sum_j A_{ij}} \sum_j A_{ij} \hat{V}_j \right\|_2,
\end{equation}
where $A_{ij}$ denotes the adjacency matrix of the facial mesh: $A_{ij} = 1$ if vertices $i$ and $j$ are connected by an edge, and $A_{ij} = 0$ otherwise. This term enforces local smoothness of the mesh surface, thereby reducing visual artifacts.

\begin{figure}[!t]
  \centering
  \includegraphics[width=\textwidth]{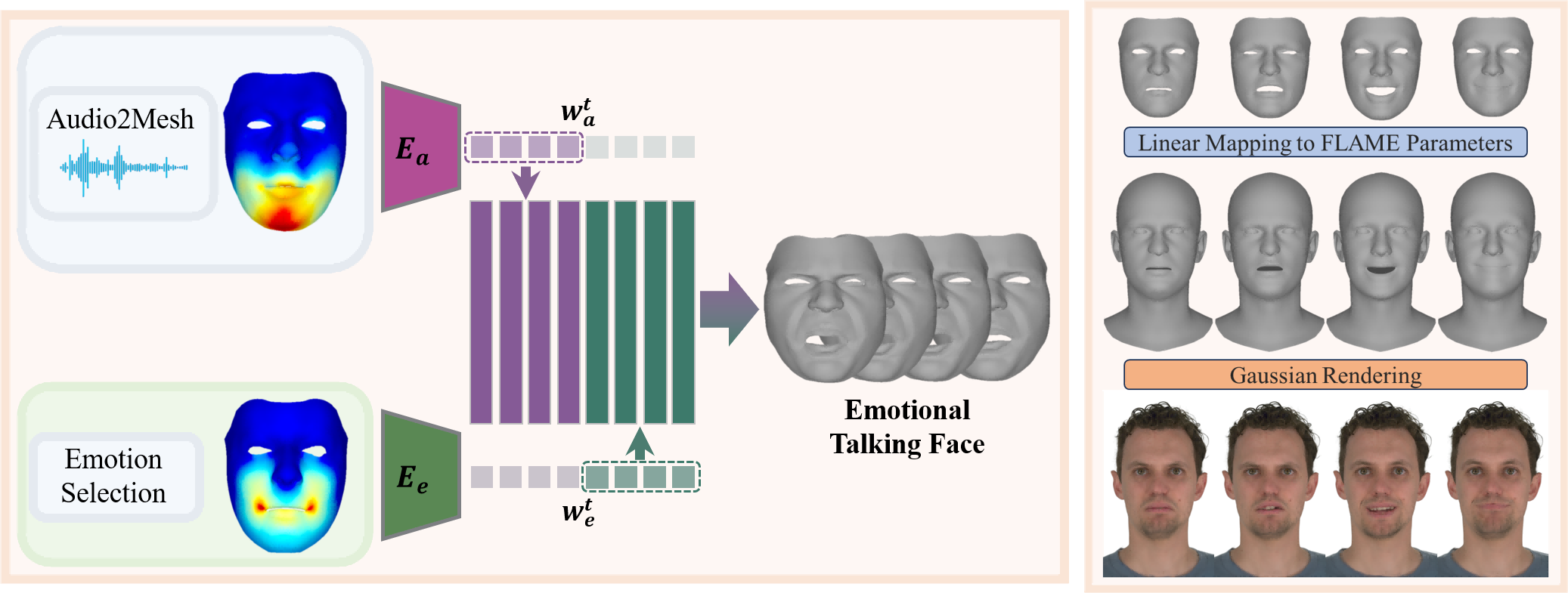}
   \caption{\textbf{Left}: Given the speech and expression deformations as input, we combine their respective blendshape weights to generate talking-face animations with expressions. \textbf{Right}: The learned blendshapes can be linearly mapped to the FLAME parameter space, enabling the animation of Gaussian head avatars.}
  \label{fig:infer}
  \vspace{-8pt}
\end{figure}

\subsection{3D Emotional Talking Face Generation}

During inference (Fig.~\ref{fig:infer}, left), we generate 3D emotional talking faces by combining the speech blendshape weights $\vec{w_{A_0}}$ and the expression blendshape weights $\vec{w_{E_0}}$.  
Given the input speech deformation $\Delta V_{A_0}$ and expression deformation $\Delta V_{E_0}$, the fused deformation is computed as:
\begin{equation}
\Delta V_{O_0} = B_A \vec{w_{A_0}} + B_E \vec{w_{E_0}}.
\end{equation}
This per-vertex deformation is then added to the canonical template mesh to produce the final talking face with the specified expression.

In addition, we learn a linear mapping $\mathrm{L}_{\mathcal{G}}$ from the learned blendshapes to the FLAME model’s expression and jaw pose parameters~\cite{li2017learning} through an optimization process with the following loss:
\begin{equation}
\mathcal{L}_{\text{mapping}} = \lambda_{\text{fit}}\mathcal{L}_{\text{fit}} + \lambda_{\text{reg}}\mathcal{L}_{\text{reg}},
\end{equation}
where $\mathcal{L}_{\text{reg}}$ is an $\ell_2$ regularization loss applied to the output deformations, and:
\begin{equation}
\mathcal{L}_{\text{fit}} = \frac{1}{n_v} \sum_{k=1}^{n_v} \omega_k \| \Delta \tilde{V}_{k} - \Delta \hat{V}_{k} \|_2
\end{equation}
is the fitting loss, which minimizes the weighted mean squared error between the vertex deformations corresponding to the FLAME model’s expression and jaw pose parameters and those generated by our deformation fusion module. This allows our method to further animate Gaussian head avatars~\cite{qian2024gaussianavatars} parameterized by FLAME, producing realistic, natural, and emotionally expressive talking faces (Fig.~\ref{fig:infer}, right).

\section{Experiments}

\subsection{Implementation Details}

We train our models using the publicly available datasets VOCAset~\cite{cudeiro2019capture} and Florence4D~\cite{principi2023florence}. VOCAset contains 29 minutes of paired audio and 3D scans from 12 speakers, while Florence4D provides dynamic sequences of 3D faces covering a wide range of expressions. Specifically, the speech-driven model is trained on VOCAset, while our deformation fusion module jointly learns the speech- and expression-driven blendshapes from both VOCAset and Florence4D. We implement our models in Pytorch~\cite{paszke2019pytorch}, and employ the Adam optimizer~\cite{kingma2014adam} with a learning rate of 3$e$-4. We use a batch size of 128 and optimize the models for 4000 epochs. Additional implementation details are provided in the supplementary materials.

\begin{table}[!b]
\centering
\begin{tabular}{@{}c|c@{}}
\toprule
    Method     & LVE (mm) $\downarrow$ \\
    \midrule
    FaceFormer       & $4.9298$ \\ 
    FaceFormer\texttt{+}DFM    & $4.9319$ \\ 
    \bottomrule
\end{tabular}
\vspace{6pt}
\caption{\textbf{Quantitative evaluation results on VOCAset.} The loss in lip-sync accuracy, as measured by LVE, caused by our deformation fusion module (DFM) is negligible.}
\vspace{-12pt}
\label{tab:quanty}
\end{table}

\subsection{Quantitative Evaluation}

Since our audio-driven model adopts the FaceFormer architecture~\cite{fan2022faceformer}, we quantitatively evaluate the impact of our deformation fusion module on lip-sync accuracy. Specifically, we assess the lip vertex error (LVE), defined as the maximum $\ell_2$ error among all lip vertices, on 119 audio samples from three subjects in our VOCAset test set. We compare the lip-sync accuracy of our audio-driven model (i.e., FaceFormer retrained on the registered VOCAset training data) against the output after applying the deformation fusion module. As reported in Tab.~\ref{tab:quanty}, the LVE increases by only 0.002134~mm after the deformation fusion module, which is imperceptible to the human eye. This demonstrates that the learned speech-driven blendshapes can accurately capture the lip deformations for speech articulation.

\subsection{Inference Speed and Efficiency}

We implement our framework in Python using the PyTorch library. The trained weights of the deformation fusion module occupy 17.89~MB of storage. As shown in Tab.~\ref{tab:speed}, our full model runs at over 165~FPS on a commercial GPU (NVIDIA GeForce RTX~3090), enabling efficient real-time generation of emotionally expressive talking-face animations in scenarios such as live video conferencing.

\begin{table}[!t]
\centering
\begin{tabular}{@{}c|c@{}}
\toprule
    Method     & Driving FPS \\
    \midrule
    FaceFormer\texttt{+}DFM    & $177.62$ \\ 
    FaceFormer\texttt{+}DFM\texttt{+}$\mathrm{L}_{\mathcal{G}}$    & $168.63$ \\ 
    \bottomrule
\end{tabular}
\vspace{6pt}
\caption{\textbf{Inference Speed.} The inference speed is measured in frames per second (FPS). Our emotional talking-face animation model, which consists of the FaceFormer-based audio-driven module and our deformation fusion module (DFM), runs at 177.62~FPS on an NVIDIA GeForce RTX~3090 GPU. After incorporating the linear mapping $\mathrm{L}_{\mathcal{G}}$ from the learned blendshapes to the FLAME parameters, the model still achieves 168.63~FPS.}
\vspace{-12pt}
\label{tab:speed}
\end{table}

\subsection{Perceptual Studies}

Since it is challenging to find an appropriate metric to automatically evaluate both the synchronization between lip articulation and input speech and the quality of emotional expression, we conducted two perceptual experiments to compare our method with previous state-of-the-art approaches. We generated 3D facial mesh sequences using different methods, rendered the outputs, and then displayed them side-by-side. Twenty-five participants rated the animations on a 5-point Likert scale~\cite{likert1932technique}, evaluating both lip-sync accuracy and the quality of emotional expression. All audio–visual stimuli were randomized independently for each participant to avoid order and position bias. Each stimulus was evaluated under all compared methods, ensuring a fully balanced design across methods.

\subsubsection{Evaluation of Lip Articulation and Emotion}

The first part of this evaluation compares our approach with publicly available 3D talking-face animation methods, including VOCA~\cite{cudeiro2019capture}, FaceFormer~\cite{fan2022faceformer}, CodeTalker~\cite{xing2023codetalker}, EmoTalk~\cite{peng2023emotalk}, and EMOTE~\cite{danecek2023emotional}. For testing, we randomly select 8 neutral voice sequences from the MEAD dataset~\cite{wang2020mead}. This round of evaluation focuses on the synchrony of mouth shapes for neutral expressions. The participants rate the rendered videos from different methods, resulting in 200 paired observations (25 participants × 8 audio–visual stimuli). For each comparison between our method and another approach, we perform one-sided paired $t$-tests on the paired ratings. Specifically, for each participant–stimulus pair, we compute the score difference, and test the alternative hypothesis $H_1:\mu_d>0$, where $\mu_d$ denotes the population mean of these paired differences. A significant positive mean difference indicates that our method was consistently rated higher. The resulting $p$-values quantify the statistical significance of the observed score improvements over previous approaches, while the effect sizes (Cohen’s $d_z$) measure their practical significance. As shown in Tab.~\ref{tab:percept_neu}, our model, which employs the retrained FaceFormer as the audio-driven backbone, performs on par with FaceFormer (no statistically significant difference, $p=0.25$) and significantly outperforms VOCA, CodeTalker, EmoTalk, and EMOTE ($p<0.001$) according to one-sided paired $t$-tests. The corresponding effect sizes (Cohen’s $d_z=0.24$–$0.73$) indicate small-to-large practical improvements, confirming that our model produces perceptually more synchronized and natural lip movements.

\begin{table}[!t]
\centering
\setlength{\tabcolsep}{6pt}
\begin{tabular}{@{}lcccccc@{}}
\toprule
\multirow{2}{*}{Method} & \multirow{2}{*}{Training Set} & \multicolumn{2}{c}{MOS $\uparrow$} & 
\multicolumn{2}{c}{Significance vs.\ Ours} \\
\cmidrule(lr){3-4} \cmidrule(l){5-6}
 & & Score $\pm$ SD &  & $p$-value & Cohen’s $d_z$ \\
\midrule
VOCA~\cite{cudeiro2019capture}       & VOCA  & $3.64 \pm 0.87$  &  & $1.8\times10^{-10}$ & 0.47 \\
CodeTalker~\cite{xing2023codetalker} & VOCA  & $3.81 \pm 0.94$  &  & $4.8\times10^{-4}$  & 0.24 \\
EmoTalk~\cite{peng2023emotalk}       & RAV/HDTF & $3.78 \pm 0.95$  &  & $8.5\times10^{-5}$  & 0.27 \\
EMOTE~\cite{danecek2023emotional}    & MEAD     & $3.22 \pm 1.10$  &  & $1.7\times10^{-20}$ & 0.73 \\
\textbf{FaceFormer}~\cite{fan2022faceformer}  & VOCA  & $\mathbf{4.02 \pm 0.81}$ &  & 0.25 & 0.05 \\
\midrule
\textbf{Ours}                        & VOCA  & $\mathbf{4.06 \pm 0.82}$ &  &  --  &  -- \\
\bottomrule
\end{tabular}
\vspace{6pt}
\caption{\textbf{Perceptual study} evaluating the lip-synchronization performance of our method against state-of-the-art approaches. Subjective lip-sync quality scores (Mean Opinion Score, MOS; higher is better) are reported as mean~$\pm$~standard deviation (SD), along with one-sided paired $t$-tests comparing each approach against our method. Reported $p$-values and Cohen’s $d_z$ indicate statistical and practical significance, respectively.}
\vspace{-12pt}
\label{tab:percept_neu}
\end{table}

\begin{table}[!t]
\centering
\setlength{\tabcolsep}{2.6pt}
\begin{tabular}{@{}clcccccc@{}}
    \toprule
    Aspect & Method     & Training Dataset & \multicolumn{2}{c}{MOS $\uparrow$} & 
\multicolumn{2}{c}{Significance vs.\ Ours} \\
\cmidrule(lr){4-5} \cmidrule(l){6-7}
& & & Score $\pm$ SD &  & $p$-value & Cohen’s $d_z$ \\
    \midrule
    \multirow{3}{*}{Lip-Sync} 
    & EmoTalk~\cite{peng2023emotalk}    & RAV/HDTF     & $ 3.17\pm 1.03$  &  & $1.4\times10^{-22}$ & 0.78 \\
    & EMOTE~\cite{danecek2023emotional} & MEAD         & $ 3.00\pm 1.04$  &  & $1.0\times10^{-30}$ & 0.97 \\
    \cmidrule{2-7}
    & \textbf{Ours}                     & VOCA\&Flr-4D & \boldmath{$ 4.08\pm 0.84$} &  &  --  &  -- \\
    \midrule
    \multirow{3}{*}{Expression}
    & EmoTalk~\cite{peng2023emotalk}    & RAV/HDTF     & $2.71\pm 1.09$  &  & $1.0\times10^{-38}$ & 1.15 \\
    & EMOTE~\cite{danecek2023emotional} & MEAD         & $ 2.70\pm 1.05$  &  & $5.9\times10^{-37}$ & 1.11 \\
    \cmidrule{2-7}
    & \textbf{Ours}                     & VOCA\&Flr-4D & \boldmath{$ 4.20\pm 1.01$ } &  &  --  &  -- \\
    \bottomrule
\end{tabular}
\vspace{6pt}
\caption{\textbf{Perceptual study} evaluating the lip-synchronization and expression quality of our method against state-of-the-art emotional talking face approaches. Mean Opinion Scores (MOS) are reported as mean~$\pm$~standard deviation (SD) for both aspects, along with one-sided paired $t$-tests comparing each approach against our method.}
\vspace{-12pt}
\label{tab:percept_emo}
\end{table}

The second part of the evaluation compares the proposed approach with publicly available emotional 3D talking-face animation methods, including EmoTalk~\cite{peng2023emotalk} and EMOTE~\cite{danecek2023emotional}. We randomly select 8 audio sequences from the MEAD dataset, each accompanied by exemplar videos as emotion references (e.g., happiness, sadness, anger, and disgust). This round of evaluation assesses both lip articulation and emotional expressiveness. The participants rate the rendered videos from different methods, resulting in 200 paired observations (25 participants × 8 audio–visual stimuli) for each aspect. As in the first part, we perform one-sided paired $t$-tests on the paired ratings to examine whether our method is rated significantly higher than the previous methods. As shown in Tab.~\ref{tab:percept_emo}, our model outperforms previous emotional 3D talking-face animation methods by approximately 1 MOS point in both evaluation aspects. One-sided paired $t$-tests show that the improvements over EmoTalk and EMOTE are statistically significant ($p < 10^{-22}$ and $p < 10^{-30}$ for lip-sync, $p < 10^{-37}$ for expression), with large effect sizes (Cohen’s $d_z = 0.78$–$1.15$). These results demonstrate that our approach achieves a substantial perceptual advantage in the mixed driving of speech and emotion. This improvement is attributed to our explicitly disentangled speech- and expression-driven blendshapes, together with the sparsity constraint loss, which reduces cross-domain interference and preserves both lip-sync accuracy and emotional expressiveness.

\begin{table}[!b]
\centering
\setlength{\tabcolsep}{6pt}
\begin{tabular}{lccc}
    \toprule 
    Configurations      & MOS (Lip-Sync)$\uparrow$ && MOS (Expression)$\uparrow$ \\
    \midrule 
    w/o $\mathcal{L}_{\text{Laplace}}$  & $3.38\pm 0.72$ && $3.52\pm 0.85$     \\
    w/o $\mathcal{L}_{\text{sparsity}}$ & $2.90\pm 0.87$ && $2.67\pm 1.21$     \\
    w/o Geometric Mapping   & $2.95\pm 0.98$     && $2.90\pm 1.06$             \\
    \midrule 
    Interpolation of Inputs & $2.71\pm 1.08$     && $2.76\pm 0.97$             \\
    \midrule 
    Ours             & \boldmath{$4.57\pm 0.58$} && \boldmath{$4.19\pm 0.66$}  \\
    \bottomrule 
\end{tabular}
\vspace{6pt}
\caption{\textbf{Ablation study} results. The full model achieves the highest scores in both lip synchronization and expression quality.}
\vspace{-12pt}
\label{table:ablation}
\end{table}

\begin{figure}[!b]
\centering
\includegraphics[width=\textwidth]{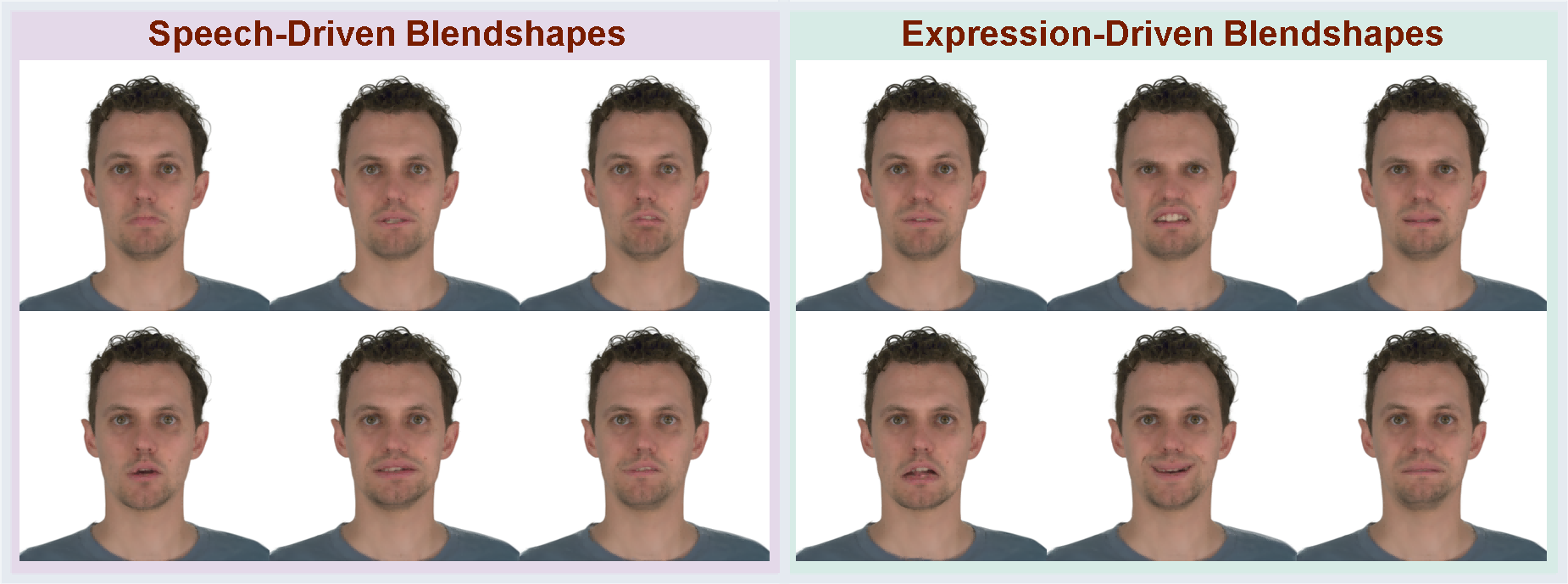}
\caption{Visualization of \textbf{learned blendshapes}. The blendshapes are linearly mapped to the FLAME parameter space and used to animate Gaussian head avatars for intuitive visualization.}
\label{fig:blendshape}
\vspace{-8pt}
\end{figure}

\subsubsection{Ablation Study: Perceptual Evaluation}

\begin{figure}[!b]
\centering
\vspace{-5pt}
\includegraphics[width=0.9\textwidth]{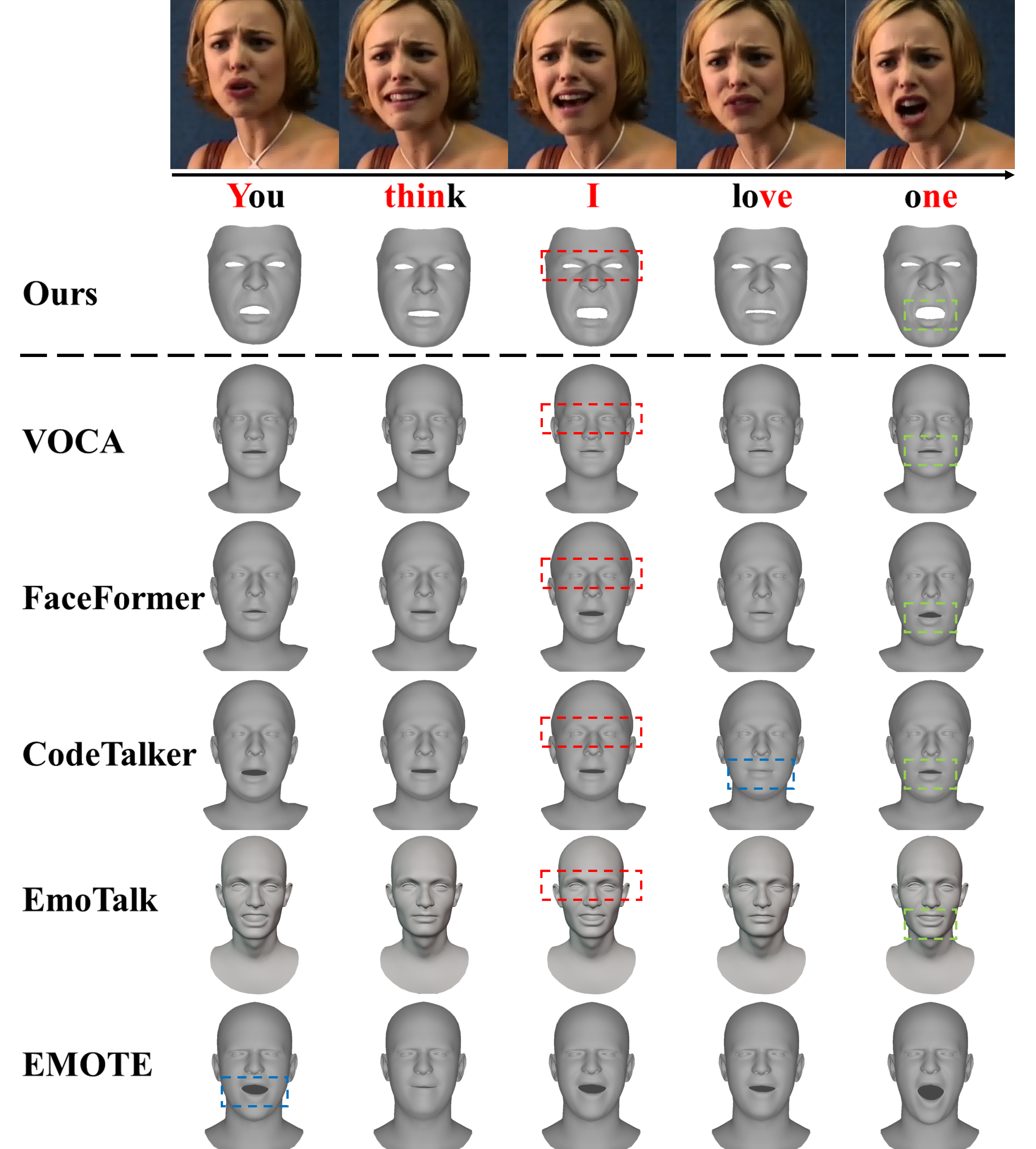}
\caption{\textbf{Qualitative Comparison with Existing Methods.} For reference, the first row shows the corresponding RGB frames. The subsequent rows present corresponding frames synthesized by our method and previous state-of-the-art methods (VOCA~\cite{cudeiro2019capture}, FaceFormer~\cite{fan2022faceformer}, CodeTalker~\cite{xing2023codetalker}, EmoTalk~\cite{peng2023emotalk}, and EMOTE~\cite{danecek2023emotional}) using their official pre-trained models.}
\label{fig:qualitative}
\vspace{-8pt}
\end{figure}

This study aims to assess the contribution of each component within our model. We conducted comparisons among: (1) without Laplace smoothing loss, (2) without sparsity constraint loss, (3) without geometric mapping, (4) with the interpolation method, where the output of our audio-driven model is directly interpolated with expressions, and (5) our full model. The experimental setup is identical to that used in the perceptual study for 3D emotional talking-face animation evaluation. The results (Tab.~\ref{table:ablation}) demonstrate the necessity of each component in our method. In particular, our approach significantly outperforms the direct interpolation of speech-driven and expression-driven deformations, indicating that our deformation fusion module can effectively regress and remove cross-domain deformations present in the inputs. Furthermore, the significant performance drop without the sparsity constraint loss validates the effectiveness of our soft disentanglement strategy during blendshape learning, which allows the model to capture minor cross-domain deformations inherent in the training data.

\subsection{Qualitative Evaluation}

\subsubsection{Visualization of Learned Blendshapes}

We visualize a subset of the speech- and expression-driven blendshapes learned in the deformation fusion module. For more intuitive presentation, the blendshapes are linearly mapped to the FLAME parameter space to animate Gaussian head avatars~\cite{qian2024gaussianavatars}. As shown in Fig.~\ref{fig:blendshape}, our model learns disentangled speech-driven and expression-driven deformations. The learned speech-driven blendshapes primarily focus on variations in mouth articulation, while the expression-driven blendshapes capture distinct emotional expressions.

\begin{figure}[!b]
\centering
\includegraphics[width=0.8\textwidth]{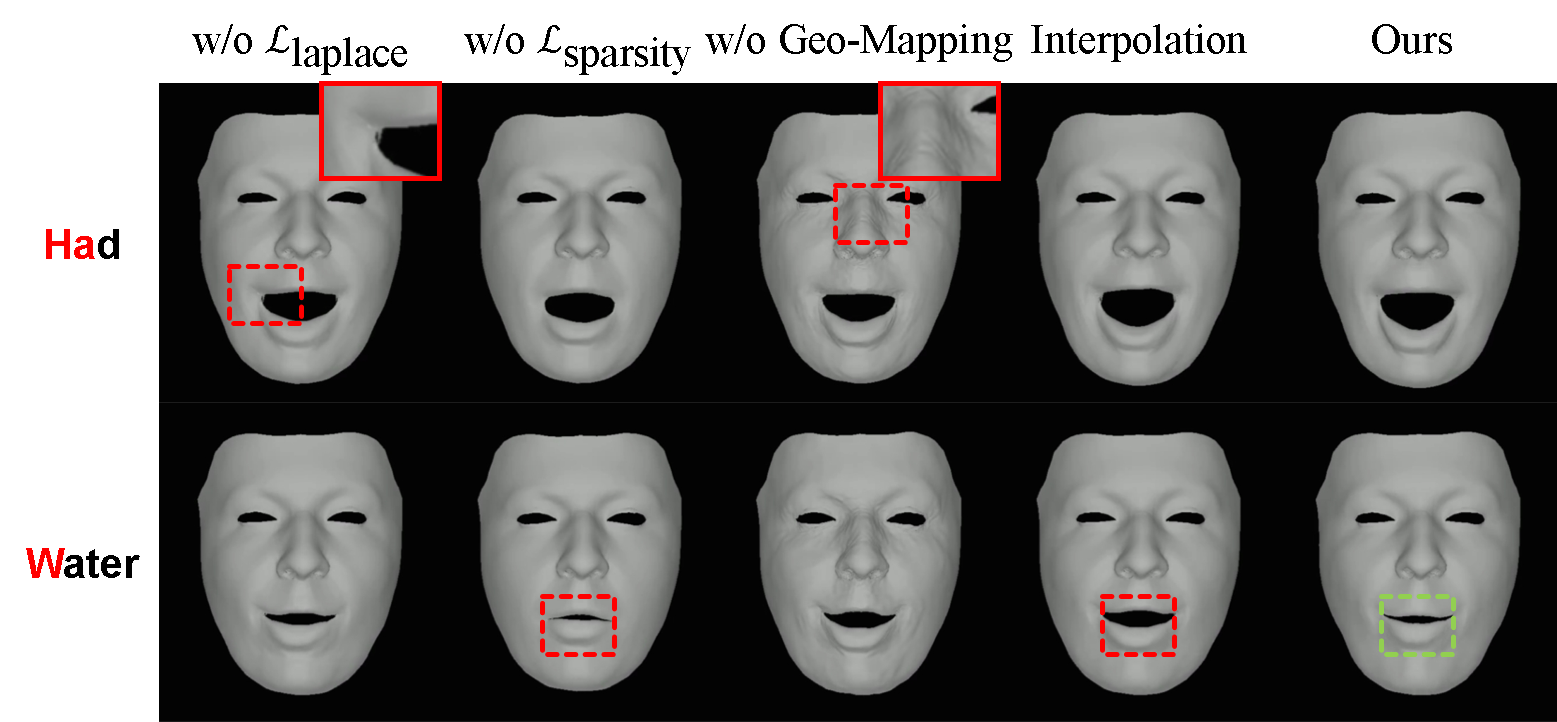}
\caption{\textbf{Visual ablation study results.} From left to right: (1) without Laplace smoothing loss, (2) without sparsity constraint loss, (3) without geometric mapping, (4) with the interpolation method, where the speech- and expression-driven deformations are directly interpolated, and (5) our full model.}
\label{fig:ablat}
\vspace{-8pt}
\end{figure}

\subsubsection{Visual Comparison}
In Fig.~\ref{fig:qualitative}, we compare our method with state-of-the-art approaches~\cite{cudeiro2019capture,fan2022faceformer,xing2023codetalker,peng2023emotalk,danecek2023emotional}. While most methods produce natural lip movements, they sometimes fail to align perfectly with the ground truth. For example, CodeTalker and EMOTE occasionally struggle to synchronize lip movements accurately with verbal signals on certain syllables (indicated by the blue dashed lines). In addition, except for EMOTE, other methods exhibit less effective performance in the upper facial region compared to ours (indicated by the red dashed lines). In the last column, while most methods are able to reflect lip movements matching the speech, they seldom account for emotional expression, as indicated by the green dashed lines. This highlights that our fusion strategy can generate high-quality facial animations in both speech and expression domains.

\subsubsection{Ablation Study: Visual Comparison}
As shown in Fig.~\ref{fig:ablat}, the model without the Laplace smoothing loss exhibits spiky artifacts around the mouth corners due to the absence of local smoothness constraints. The model without the sparsity-constrained loss fails to produce correct emotional expressions, as speech- and expression-driven deformations are not disentangled during training. The model without geometric mapping produces uneven artifacts because it cannot leverage the locality of 3D mesh deformations through the CNN architecture. Finally, directly interpolating speech- and expression-driven deformations disregards the inherent secondary cross-domain deformations in the inputs, which can cause issues such as the lips failing to close properly for phonemes like /w/ due to the influence of a smiling expression. This supports our hypothesis that cross-domain deformations should be regressed and removed during the fusion of speech- and expression-driven deformations.

\section{Conclusion}
We presented a novel approach for 3D emotional talking-face animation that learns disentangled speech-driven and expression-driven deformations. By modeling facial animation as a linear combination of two blendshape sets learned from high-quality 3D scans, our method captures realistic lip motions and rich emotional expressions. Our main contribution lies in the deformation fusion module trained with the sparsity constraint loss, which effectively separates the two driving factors while accommodating subtle cross-domain effects. Experiments demonstrate that our approach improves emotional expressiveness without sacrificing speech articulation accuracy. Furthermore, the learned blendshape parameters can be mapped to the FLAME model, enabling the animation of high-quality Gaussian head avatars.

We present a novel approach for 3D emotional talking-face animation that learns disentangled speech-driven and expression-driven deformations. By modeling facial animation as a linear combination of two blendshape sets learned from high-quality 3D scans, our method effectively captures both realistic lip motions and rich emotional expressions. Our main contribution lies in the deformation fusion module trained with the sparsity constraint loss, which effectively separates the two driving factors while accommodating subtle cross-domain effects. Experiments demonstrate that our approach improves emotional expressiveness without sacrificing speech articulation accuracy. Our model is lightweight and achieves a frame rate of over 165 FPS on a commercial GPU (NVIDIA GeForce RTX~3090). Furthermore, the learned blendshape parameters can be mapped to the FLAME model, enabling the animation of high-quality Gaussian head avatars. This allows for the generation of more realistic, emotionally expressive talking faces, with potential applications in real-time interactive environments such as XR, where improved expression-speech alignment can enhance user engagement, social presence, and immersion.

\bibliographystyle{splncs04}
\bibliography{ref}






\end{document}